\documentclass[letterpaper, 10pt, conference]{ieeeconf}
\IEEEoverridecommandlockouts
\usepackage[utf8]{inputenc}
\usepackage{amsmath}
\usepackage{amssymb}
\usepackage{color}
\usepackage{graphicx}
\usepackage[pdfencoding=auto]{hyperref}
\usepackage{bm}
\usepackage{graphicx}
\usepackage{comment}
\usepackage{soul}
\usepackage{pdfpages}
\usepackage{subfigure}
\sethlcolor{green}

\newcommand{\defeq}{\stackrel{\mathrm{def}}{=}}

\def\bfcd{\dot{\bfc}}

\def\bfqd{\dot{\bfq}}

\def\bfcdd{\ddot{\bfc}}

\def\bfqdd{\ddot{\bfq}}

\newcommand{\bftau}{\boldsymbol{\tau}}

\newcommand{\bfUpsilon}{\boldsymbol{\Upsilon}}

\newcommand{\bfPsi}{\boldsymbol{\Psi}}

\newcommand{\bfa}{\ensuremath {\bm{a}}}
\newcommand{\bfb}{\ensuremath {\bm{b}}}
\newcommand{\bfc}{\ensuremath {\bm{c}}}

\newcommand{\bfe}{\ensuremath {\bm{e}}}
\newcommand{\bff}{\ensuremath {\bm{f}}}
\newcommand{\bfg}{\ensuremath {\bm{g}}}

\newcommand{\bfp}{\ensuremath {\bm{p}}}
\newcommand{\bfq}{\ensuremath {\bm{q}}}

\newcommand{\bfw}{\ensuremath {\bm{w}}}

\newcommand{\bfA}{\mathbf{A}}

\newcommand{\bfE}{\mathbf{E}}
\newcommand{\bfF}{\mathbf{F}}
\newcommand{\bfG}{\mathbf{G}}

\newcommand{\bfI}{\mathbf{I}}
\newcommand{\bfJ}{\mathbf{J}}
\newcommand{\bfK}{\mathbf{K}}

\newcommand{\bfP}{\mathbf{P}}

\newcommand{\bfS}{\mathbf{S}}

\newcommand{\bfW}{\mathbf{W}}

\newcommand{\bfY}{\mathbf{Y}}

\newcommand{\calC}{{\cal C}}

\title{\Large \bf
	Balance of Humanoid robot in Multi-contact and Sliding Scenarios}

\author{Saeid Samadi, St\'{e}phane Caron, Arnaud Tanguy, and Abderrahmane Kheddar
	\thanks{The authors are with the Montpellier Laboratory of Informatics,
		Robotics and Microelectronics (LIRMM), CNRS-University of Montpellier,
		France. A. Kheddar is also with the CNRS-AIST Joint Robotics Laboratory
		(JRL), UMI3218/RL, Tsukuba, Japan. 
		Corresponding author: {\tt\footnotesize saeid.samadi@lirmm.fr}}%
}

\begin{document}
	
	\maketitle
	\thispagestyle{empty}
	\pagestyle{empty}
	
	\begin{abstract}
		This study deals with the balance of humanoid or multi-legged robots in a multi-contact setting where a chosen subset of contacts is undergoing desired sliding-task motions. One method to keep balance is to hold the center-of-mass (CoM) within an admissible convex area. This area should be calculated based on the contact positions and forces. We introduce a methodology to compute this CoM support area (CSA) for multiple fixed and sliding contacts. To select the most appropriate CoM position inside CSA, we account for (i) constraints of multiple fixed and sliding contacts, (ii) desired wrench distribution for contacts, and (iii) desired position of CoM (eventually dictated by other tasks). These are formulated as a quadratic programming optimization problem. We illustrate our approach with pushing against a wall and wiping and conducted experiments using the HRP-4 humanoid robot.
	\end{abstract}
	
	
	\section{Introduction}
	
	Humanoid robots are designed to interact with the environment and are supposed to be able to reproduce all sorts of physical actions that a human could do such as running, flipping, jumping, crawling, etc. In theory, state-of-the-art humanoid robots have the hardware capability to achieve --to some extended and relative performances, some of these complex behaviors. Yet, they lack efficient control strategies with robust equilibrium conditions.

	Sustaining equilibrium is more challenging in multi-contact settings. Balance in multi-contact conditions was studied theoretically in~\cite{Collette2007} but restricted to non-sliding contacts. A recent study on the interaction of the robot with the environment introduces a passivity-based whole body balancing framework~\cite{Abi-Farraj2019}. In this latter work, gravito-Inertial wrench cone (GIWC) --introduced in~\cite{Caron2015icra}, is used to keep the balance in multi-contact, eventually under moving but not sliding contacts. 
	
	Sliding motion is one of the actions that humans master in achieving several tasks in their daily lives. Sliding contacts have been studied on several robots by considering dynamic friction forces and planning and controlling objects by pushing, see a recent example in~\cite{shi2017tro}. 
	
	Recent works on humanoid robots are mostly focused on avoiding~\cite{Kajita2004,Zhou2018} or recovering~\cite{Kaneko2005,Vazquez2013} slips. Keeping the balance while sliding on purpose has been studied for two feet contacts in both rotational and translational directions. Linear contact constraints have been used in~\cite{Kojima2017} for the controller to keep the dynamic balance while slipping in rotational directions on the feet. While rotational-slipping, considering the force distribution within the contact surfaces is the most challenging part of the study. Slip-turn motion for the robot is produced in~\cite{Miura2013} by minimizing the floor friction power. The works in~\cite{Hashimoto2011,Koeda2011} are about generating quick turning motion by rotational shuffling.
	\begin{figure}[!tb]
		\centering 
		\includegraphics[width=\columnwidth]{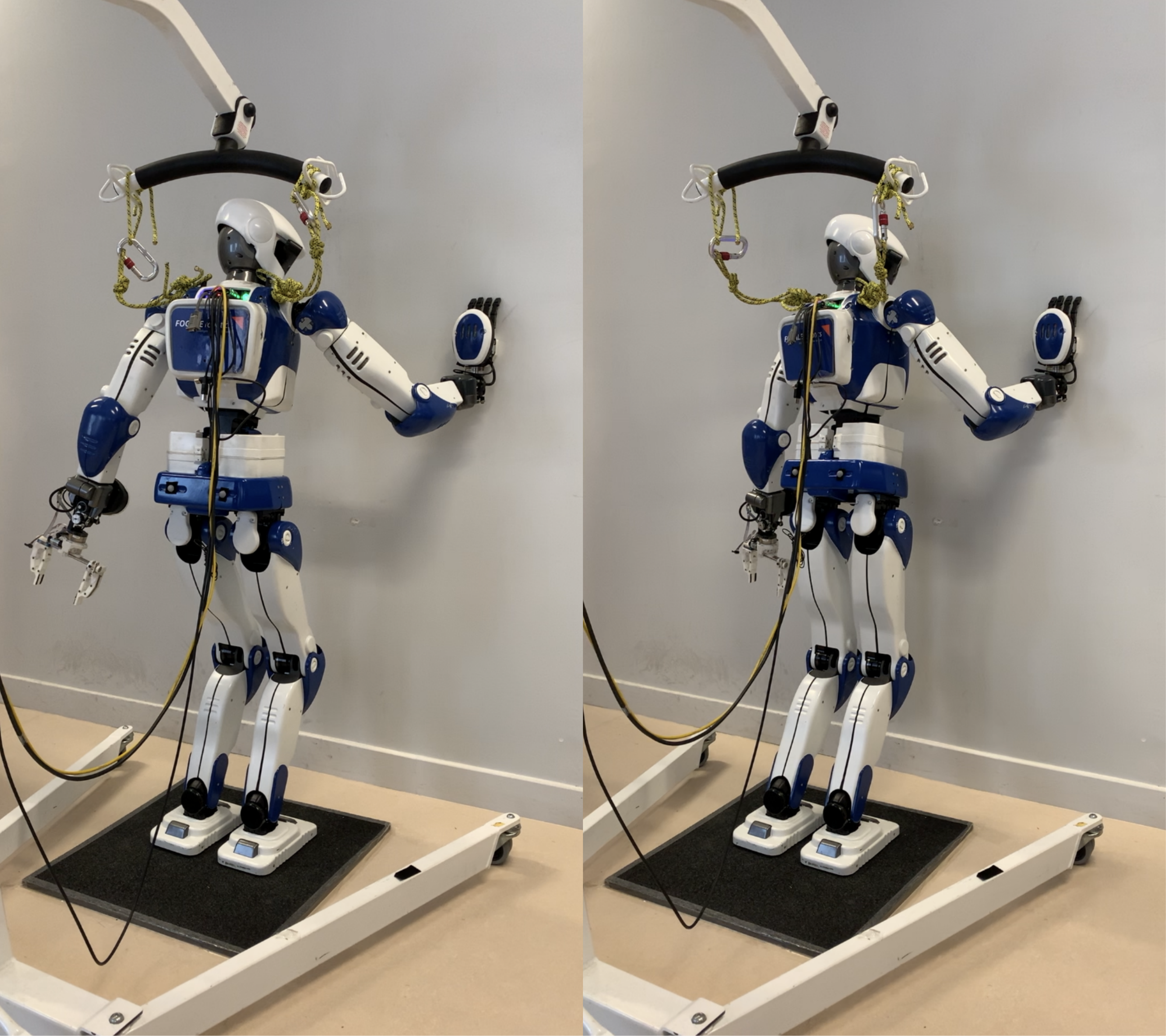}
		\caption{Keeping balance of HRP-4 humanoid robot in pushing-against-a-wall scenario by adjusting position of the CoM.}
		\label{firstpic}
	\end{figure}

	As a motion planning for translational shuffling, \cite{Kojima2015} implemented two-layer controller to satisfy of the stability criteria of the robot using Zero-tilting Moment Point (ZMP) and achieve proper force distribution in two feet contacts. Total motion sequence of this method consists of both slide-and-stop phases and CoM trajectory is generated solving a QP and related constraints. Also, keeping the dynamic balance while shuffling is studied in~\cite{Posa2014} by formulating complementarity constraints into a QP, and in~\cite{Kim2011} by uniform force distribution assumption on the sole for single support phase.

	Multiple sliding contacts for humanoid robots are still a challenge to solve. Our contribution to this field of interest is in introducing an applicable method to keep the balance of the robot in multi-contact settings. Our method not only guarantees equilibrium criteria for fixed contacts, as shown in Figure~\eqref{firstpic}, but also keeps the balance of the robot in the presence of sliding contacts.
	
	In the following, we structured our work in four sections. In Section~\ref{CSA}, we introduce a CoM support area for multiple fixed and sliding contacts. Holding the position of the CoM inside this area guarantees the balance of the robot while some chosen contacts slide and other not. Next, in Section~\ref{CQP}, we illustrate our methodology for calculating the position of CoM under constraints. Sections~\ref{exp} and~\ref{conc} shows the experimental results of proposed method and gives a conclusion of the whole work, respectively.

	\section{CoM support area} \label{CSA}
	
	The dynamic model of any robot which can be described by Newton-Euler equations is in the following form:
	\begin{equation}
	\sum {\bfW_c} = - \bfW_m \label{NE}
	\end{equation}
	where $\bfW_c \in \mathbb{R}^6$ and $\bfW_m\in \mathbb{R}^6$ are the contact and the gravity wrenches respectively. We follow the common practice to write the resultant linear term (force) first in the wrench followed by the moment term. So, $ \mathbf{W} = [\mathbf{f} \ \ \mathbf{\bftau} ]^T$. By separating forces and torques of~\eqref{NE}, we have:
	\begin{equation}
	\sum_i \bff_i=-m\bfg \label{NE1}
	\end{equation}
	\begin{equation}
	\sum_i {\bfp_i\times \bff_i}=-\mathbf{p}_G\times{m\bfg} \label{NE2}
	\end{equation}
	where $m$ is the total mass of the robot,
	$\bfp_i , \bff_i \in \mathbb{R}^3$ are the position and force of the $i^{\text{th}}$ contact point respectively;
	$\bfg\in \mathbb{R}^3$ is the gravity vector and is equal to $\begin{bmatrix}
	0 &  0 & -9.81
	\end{bmatrix}^\text{T}$ and 
	$\mathbf{p}_G \in \mathbb{R}^3$ is the position of the CoM.
	By introducing the unit vector $\textbf{e}_z$ as $\begin{bmatrix}
	0 &  0 & 1
	\end{bmatrix}^T$, we can rewrite these equations by separating the vertical component of the gravity vector:
	\begin{equation}
	\sum_i {\bff_i}=mg\bfe_z \label{meq1}
	\end{equation}
	\begin{equation}
	\sum_i {\bfp_i\times{\bff_i}}=mg\mathbf{p}_G\times{\bfe_z} \label{meq2}
	\end{equation}
	where gravity acceleration is $g = -9.81$~m/{s$^2$. Moreover, by applying a cross-product to both sides of~\eqref{meq2} by $\textbf{e}_z$, we get the following equation:
		\begin{equation}
		mg\bfe_z\times \mathbf{p}_G \times \bfe_z=\sum_i {\bfe_z\times(\bfp_i\times \bff_i)} \label{crossP0}
		\end{equation}
		
		Next, we use two well-known properties of the cross-product that are:
		\begin{equation}
		\bfe_z\times \mathbf{p}_G \times \bfe_z=\begin{pmatrix}
		\mathbf{p}_G^x\\ \mathbf{p}_G^y\\0
		\end{pmatrix} \label{crossP1}
		\end{equation}
		
		\begin{equation}
		\bfa\times (\bfb\times \bfc)=(\bfa.\bfc)\bfb-(\bfa.\bfb)\bfc \label{crossP2}
		\end{equation}
		by applying equations~\ref{crossP1} and~\ref{crossP2} into~\eqref{crossP0} we have:
		\begin{equation}
		mg\mathbf{p}_G^{S}=\sum_i {f_i^z \bfp_i-p_i^z \bff_i} \label{NE5}
		\end{equation}
		where $\mathbf{p}_G^{S}$ denotes the position of the CoM in the horizontal plane and is equal to $[p_G^x \ p_G^y \ 0]^T$. 
		
		Let's consider we have a robot in a 3-contacts setting. The position of the CoM will be formulated as follows:
		\begin{equation}
		\begin{split}
		\mathbf{p}_G^{S}=&\frac{f_3^z}{mg}\bfp_3+\frac{f_2^z}{mg}\bfp_2+\frac{f_1^z}{mg}\bfp_1 \\
		&-[\frac{p_3^z}{mg}\bff_3+\frac{p_2^z}{mg}\bff_2+\frac{p_1^z}{mg}\bff_1]
		\end{split} \label{CoMPos1}
		\end{equation}
		
		The equation~\ref{CoMPos1} shows the general form of this formula regardless of orientation and positions of contacts. As a particular case, we assume that both feet are on the ground so that  $p_1^z=p_2^z=0$. Consequently~\eqref{CoMPos1} becomes:
		\begin{equation}
		\mathbf{p}_G^{S}=\frac{f_1^z}{mg}\bfp_1+\frac{f_2^z}{mg}\bfp_2+\frac{f_3^z}{mg}\bfp_3-\frac{p_3^z}{mg}\bff_3 \label{meq6}
		\end{equation}
		
		In the following, we specify a region for the feasible position of CoM without losing the balance according to our sliding and multi-contact conditions.
		This region was introduced in~\cite{Bretl2008} as a static equilibrium CoM area, and its extension to 3D in~\cite{audren2018tro}.
		The purpose of our present work is to be able to balance the robot in multi-contact configurations with fixed or sliding contacts. Furthermore, this method will guarantee the stability of the robot for sliding contacts such as wiping a board by hand or shuffling foot motions.
		
		Let's consider a humanoid having its feet on the ground and one of its arm (e.g. right one) wiping a board (non-coplanar with the other contacts). The wiping trajectory could be anything and there is no limitation for the direction of wiping.
		Hence, the position of the sliding contact is a pre-defined parameter according to the designed wiping trajectory. 
		Therefore, the last two elements of~\eqref{meq6} is not related to the main variables ($p_1$ and $p_2$) for introducing the CoM support Area (CSA), and we can define them as an independent variable $\bfA$:
		\begin{subequations}
			\begin{align}
			\mathbf{p}_G^{S} &=\frac{f_1^z}{mg}\bfp_1+\frac{f_2^z}{mg}\bfp_2+\bfA \label{meq7} \\
			\bfA &= \frac{f_3^z}{mg}\bfp_3-\frac{p_3^z}{mg}\bff_3 \label{Adef}
			\end{align}
		\end{subequations}
		then:
		\begin{equation}
		\mathbf{p}_G^{S}-\bfA=\frac{f_1^z}{mg}\bfp_1+\frac{f_2^z}{mg}\bfp_2  \label{meq8}
		\end{equation}
		
		On the other hand, by considering~\eqref{NE1} in vertical direction, we have: 
		\begin{equation}
		\frac{f_1^z}{mg}+\frac{f_2^z}{mg}+\frac{f_3^z}{mg}=1 \label{meq4}
		\end{equation}
		
		From~\eqref{meq4}, we introduce ``$S_c$'' as the sum of the coefficients of~\eqref{meq7}:
		\begin{equation}
		S_c \defeq 1-\frac{f_3^z}{mg} 
		\end{equation}
		and \eqref{meq8} becomes:
		\begin{subequations}
			\begin{align}
			\mathbf{p}_G^{S} - \bfA &= \sum{\alpha_i(S_c\bfp_i)} \label{meq10} \\
			\alpha_i &= \frac{f_i}{S_cmg}
			\end{align}
		\end{subequations}
		where,
		\begin{equation}
		\sum \alpha_i = 1 \label{cvxhullcondition}
		\end{equation}
		Equation~\eqref{cvxhullcondition} is a sufficient condition to show that the point $\mathbf{p}_G^{S} - \bfA$ should be inside the convex polygon constructed by connecting $\bfp_i$ points.
		There are two equivalent ways to represent the wrench applied by the environment on the robot under a surface contact:
		\begin{enumerate}
			\item Contact forces applied at the vertices of the contact area~\cite{Caron2015icra};
			\item A single contact wrench applied at a given point~\cite{audren2018tro}.
		\end{enumerate}
		
		We choose the latter method.
		To be able to use the surface contact instead of single points, we replace each of these points ($\bfp_i$) with four edges of the related foot.
		Coordinates of new points are available by considering the dimension of each foot.
		CSA is depicted in Fig.~\ref{CSAfig}. The green area on the ground of Fig.~\ref{Choreonoid} is CSA for close foot contacts and Fig.~\ref{pymanoid} emphasizes this area for far feet contacts in wiping motion.
		\setcounter{footnote}{\value{footnote} - 2}  
		\begin{figure}[!tb] 
			\subfigure[]{
				\includegraphics[height=0.35\columnwidth]{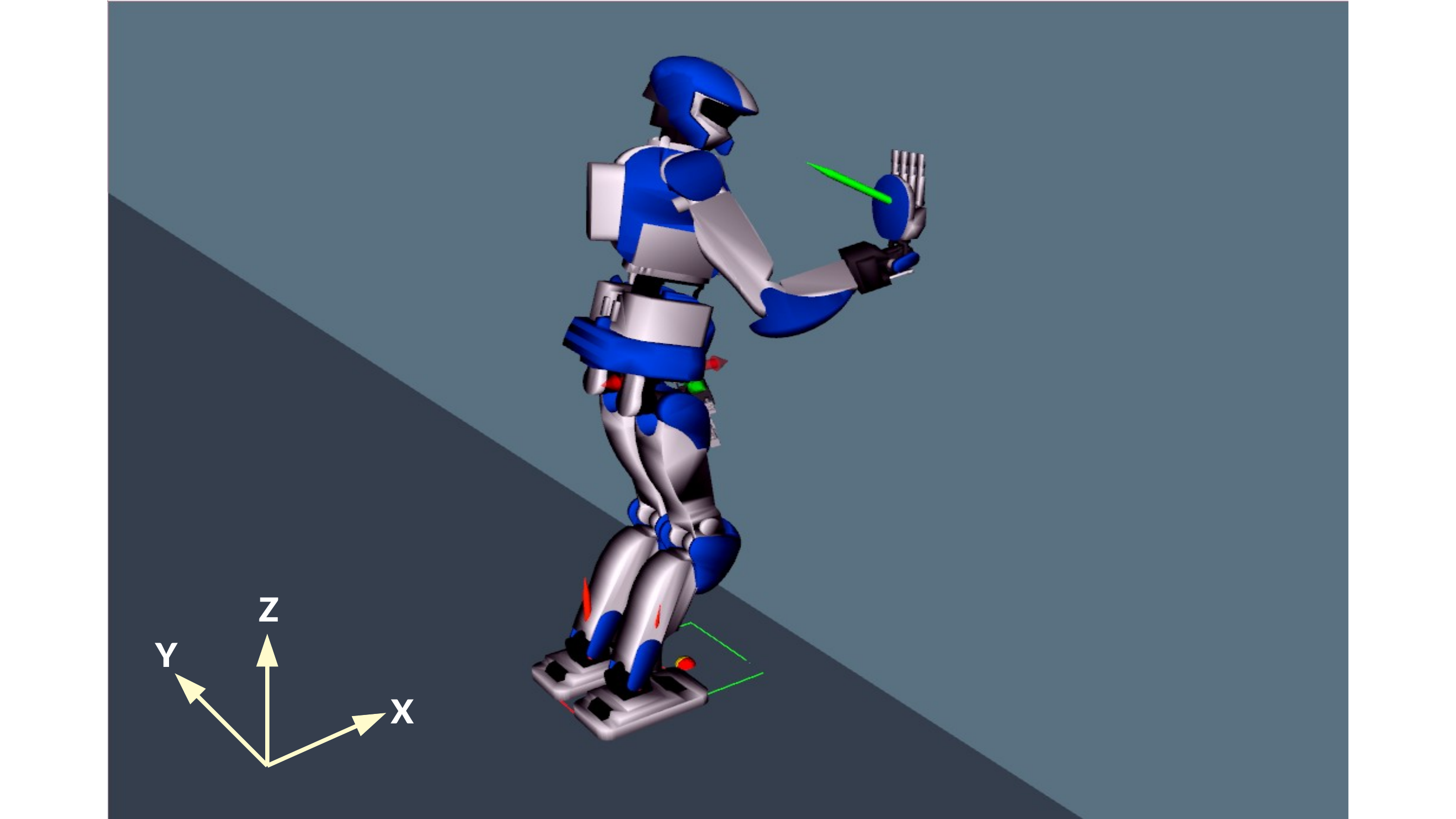}
				\label{Choreonoid}
			} 
			\subfigure[]{
				\includegraphics[height=0.35\columnwidth]{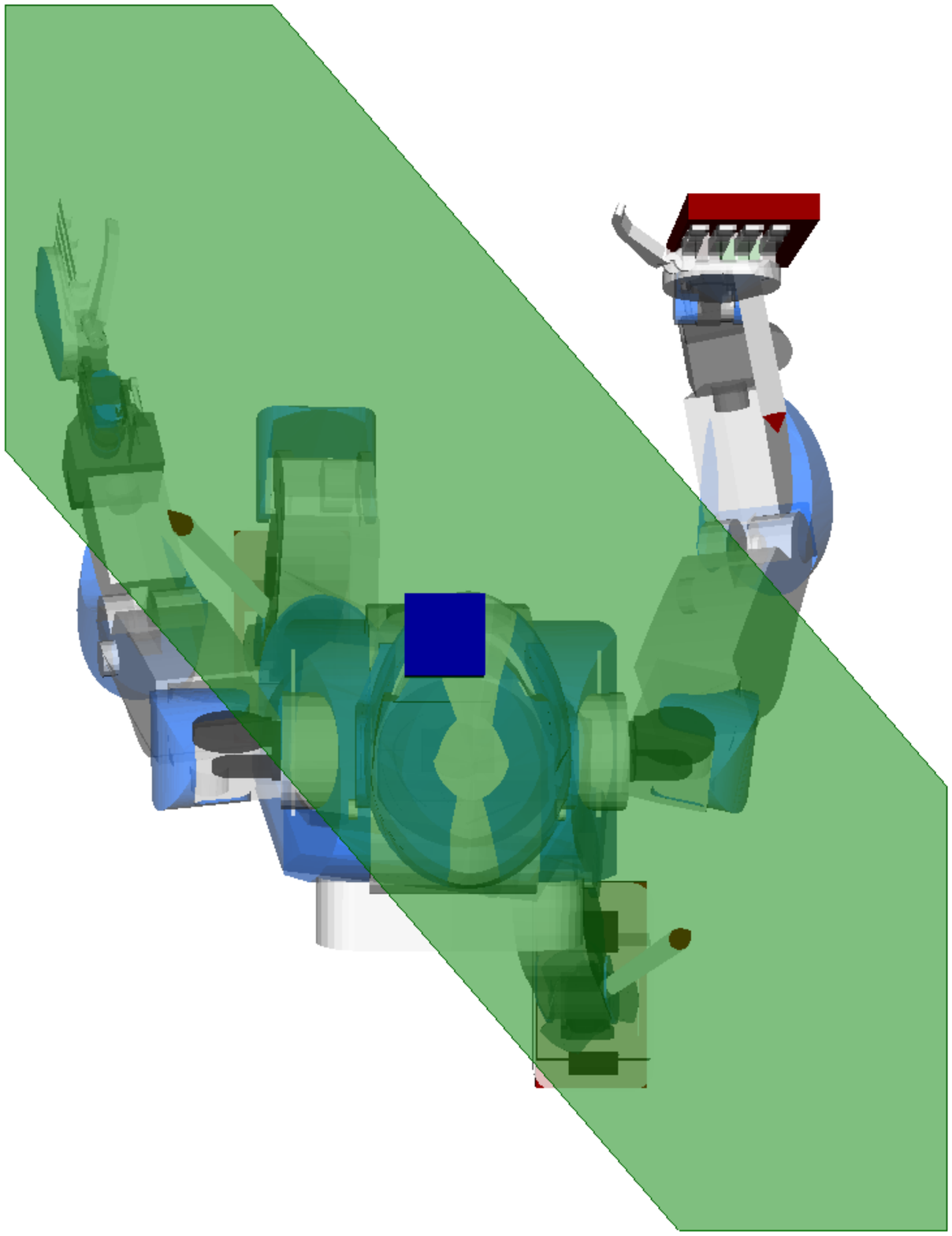} \label{pymanoid}
			} 
			\caption{Humanoid robot with sliding hand contact on the wall simulated dynamically by Choroenoid\protect\footnotemark (a) and prototyped in pymanoid\protect\footnotemark (b).}
			
			\label{CSAfig}
		\end{figure}
		
		\setcounter{footnote}{\value{footnote} - 1}  
		\footnotetext{\url{http://choreonoid.org/en/}}
		\setcounter{footnote}{\value{footnote} + 1}  
		\footnotetext{\url{https://github.com/stephane-caron/pymanoid/}}
		
		As a result, the CSA is constructed according to both sliding and multi-contact conditions. To keep the balance of the robot, the CoM should remain inside the CSA during operating motions. Setting the exact position for the CoM inside the CSA is the next contribution of our work. This is done by formulating this problem as a QP where desired behaviors are specified via a set of tasks and constraints.
		
		\section{Centroidal Quadratic Program} \label{CQP}
		
		A common solution for keeping the balance of the robot is to put a constraint on the position of the CoM and hold it inside the CSA.
		This constraint will ignore some parameters that affect the performance of the robot such as keeping the maximum distance of CoM from edges of the CSA or considering an appropriate wrench distribution on the contacts.
		
		\subsection{Decision Variables} \label{wrenches-section}
		
		Collete~\emph{et. al}~\cite{Collette2007} introduced a quadratic program that takes into account to the position of CoM, linearized contact forces and the gravity wrench in static equilibrium. Here we extend this QP to enable sliding contacts and apply sliding constraints to the related contact. Our QP is designed with the following decision variables:
		\begin{equation}
		\bfY = \begin{bmatrix}
		\mathbf{p}_G^{S} & \bfW_{\text{rf}} & \bfW_{\text{lf}} & \bfW_{\text{rh}}
		\end{bmatrix}_{1 \times 21}^T \label{decisionVariables}
		\end{equation}
		where $\bfW$ denotes the wrench in the world frame and subscripts ``$\text{rf}$", ``$\text{lf}$" and ``$\text{rh}$" correspond to the right-foot, left-foot and right-hand contacts, respectively. The solver deals with position of the CoM. This centroidal QP computes decision variables at each time step and sends them to the robot as a command. The static equilibrium of the system is given in~\eqref{NE}. Accordingly, by considering three contacts (one hand contact and two feet contacts), Newton-Euler equation writes:
		\begin{equation}\bfW_{m}+\bfW_{\text{rf}}+\bfW_{\text{lf}}+\bfW_{\text{rh}} = \textbf{$0$}
		\label{wrenchEq}
		\end{equation}
		where wrenches can be formulated as follows:
		\begin{equation}
		\textbf{$\bfW_{m}$}=\begin{bmatrix}
		0 & 0 & 0\\0 & 0 & 0\\0 & 0 & 0\\
		0 & -mg & 0\\
		mg & 0 & 0\\
		0 & 0 & 0
		\end{bmatrix} \mathbf{p}_G^{S}+\begin{bmatrix}
		0\\0\\-mg\\0\\0\\0
		\end{bmatrix}
		\end{equation}

		\begin{equation}
		\textbf{$\bfW_{\text{rf},\text{lf}}$} = \begin{bmatrix}
		& & & \\ & I_{3\times 3} & & 0_{3\times 3}\\ & & & \\ 0 & -p_z & p_y &\\p_z & 0 & -p_x & I_{3\times 3}\\-p_y & 0 & p_x &\\
		\end{bmatrix}\bfw^{\text{rf},\text{lf}}
		\end{equation}
		where $\bfw$ shows the wrench in contact frame and is equal to $[\bff \in \mathbb{R}^3 \ \bftau \in \mathbb{R}^4]^T$.
		\textbf{$\bfW_{\text{rh}}$} is calculated in a same way with \textbf{$\bfW_{\text{rf},\text{lf}}$}. In this way, we can shortly write these equations as:
		\begin{subequations}\label{w}
			\begin{align}
			\textbf{$\bfW_{\text{m}}$} &= \bfE^{m1}.\mathbf{p}_G^{S} + \bfE^{m2} \label{w1}\\
			\textbf{$\bfW_{\text{rf}}$} &= \bfE^{\text{rf}}\bfW_{\text{rf}} \label{w2} \\
			\textbf{$\bfW_{\text{lf}}$} &= \bfE^{\text{lf}}\bfW_{\text{lf}} \label{w3} \\
			\textbf{$\bfW_{\text{rh}}$} &= \bfE^{\text{rh}}\bfW_{\text{rh}} \label{w4}
			\end{align} 
		\end{subequations}
		
		\subsection{Sliding Condition} \label{sliding-section}
		
		Consider a box in direct contact with the ground as shown in Fig.~\eqref{cubebox}. An external force $\bfF_{\text{ext}} $ is applied to this box. The box will not move as long as the contact force $\bff_c$ lies within the Coulomb friction cone $\calC$~\cite{Bouyarmane2017book}.
		\begin{figure}[!htb]
			\centering
			\includegraphics[height=0.35\columnwidth]{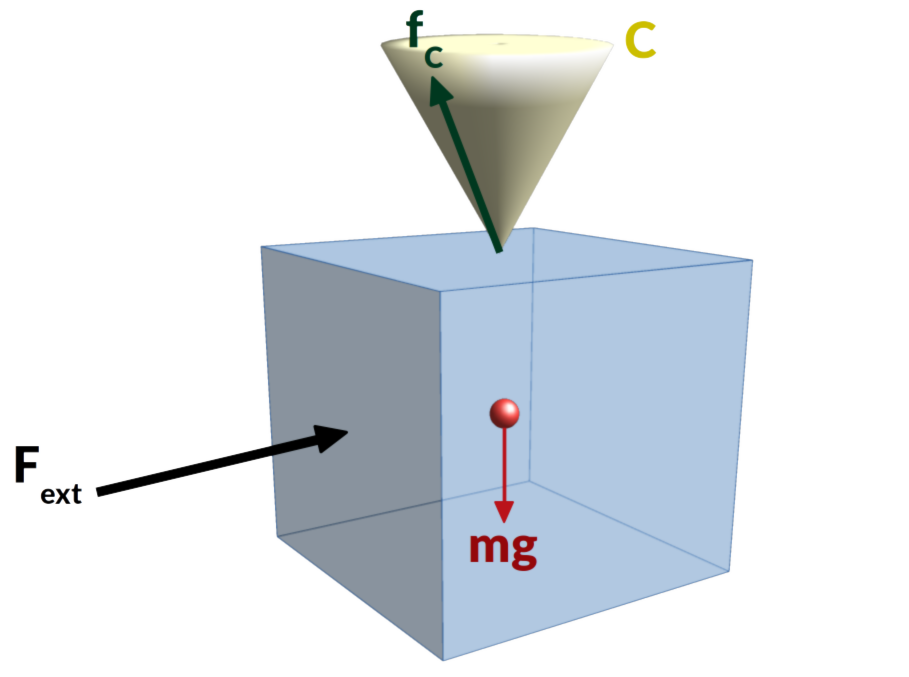} 
			\caption{Friction cone of a fixed box in presence of external force}
			\label{cubebox}
		\end{figure} 
		
		Usual works in locomotion, multi-contact conditions or motion generation for humanoids avoid slippage whereas we aim for it when needed. To avoid slipping, it is common to use inequality constraints to hold each contact force within its associated friction cone. Instead, we are implementing constraints to generate the sliding motion.
		
		For generating controlled slipping motion, force vector of the sliding contact ($\bff_c$) should remain at the edge of the related contact's friction cone, and hence expressed as equality constraints. Hand torques are considered as inequality constraints. 
		
		We introduce two matrices named $[\bfS^1]$ and $[\bfS^2]$ such that the normal hand contact force and the related friction forces could be given to the QP as a constrain:
		
		\begin{equation}
		[\bfS^1]_{6 \times 6} = 
		\begin{bmatrix}
		\textbf{I}_{3 \times 3} & \textbf{0}_{3 \times 3} \\
		\textbf{0}_{3 \times 3} & \textbf{0}_{3 \times 3} \\
		\end{bmatrix} \label{sl1}
		\end{equation}
		so that:
		\begin{equation}
		[\bfS^1] \bfW_{\text{rh}} = \begin{bmatrix}
		\bff_{3 \times 1} \\ \textbf{0}_{3\times 1}
		\end{bmatrix}_{\text{rh}} \label{sliding}
		\end{equation}

		For the sliding contact calculations, $\bff^\text{x}_{\text{rh}}$ shows the normal force applied to the right hand in the world frame. According to the wiping trajectory, the contact forces in $y$ and $z$ directions could be calculated:
		\begin{equation}
		\begin{bmatrix}
		\bff_{3 \times 1} \\ \textbf{0}_{3\times 1}
		\end{bmatrix}_{\text{rh}}  = 
		\begin{bmatrix}
		0 & 0&\dots &0\\
		\frac{f_y}{f_x} & 0&\dots &0\\
		\frac{f_z}{f_x} & \vdots & &\vdots\\  
		\vdots &  & \\
		0& 0 &\hdots &0\\
		
		\end{bmatrix}_{6 \times 6}\bfw_{\text{rh}} 
		+
		\begin{bmatrix}
		f_x  \\ 0 \\ 0 \\  
		0 \\ 0 \\0
		\end{bmatrix} \label{sliding eq}
		\end{equation}
		where $\frac{f_y}{f_x}$ and $\frac{f_z}{f_x}$ are constant numbers and they are independent from the forces. From~\eqref{sliding} and~\eqref{sliding eq}, we have:
		\begin{equation}
		[\bfS^1]\bfw_{\text{rh}} = [\bfS^2]\bfw_{\text{rh}}+\textbf{k}
		\end{equation}
		thereby:
		\begin{equation}
		[\bfS^1 - \bfS^2]\bfw_{\text{rh}} = \textbf{k} \label{sl_final}
		\end{equation}
		
		\subsection{Non-sliding Conditions} \label{Inequality Constraints}
		
		The sliding contacts are expressed with equality constraints. However, to keep the rest of the contacts fixed on the ground, we have to implement inequality constraints to maintain the related contact force inside the friction cone. The general constraints that should be applied to contact wrenches in the QP solver are same as~\cite{CaronThesis2015}:
		\begin{equation}
		\begin{split}
		\mid f_x  \mid \leqslant \mu f_z \;,\; \mid f_y \mid \leqslant \mu f_z  \;,\;  f_z^{\min} \leqslant f_z  \leqslant f_z^{\min} \\ 
		\mid \tau_x \mid  \leqslant  \mathbb{Y}f_z \;,\;  \mid \tau_y \mid \leqslant  \mathbb{X}f_z  \;,\; \tau_z^{\min} \leqslant \tau_z  \leqslant \tau_z^{\min} 
		\end{split}
		\label{boundaries}
		\end{equation}
		where $\mathbb{X}$ and $\mathbb{Y}$ are dimentions of contact surfaces, so:
		\begin{equation}
		\tau_z^{\min} =
		-\mu(\mathbb{X}+\mathbb{Y})f_z+\mid \mathbb{Y}f_x - \mu \tau_x \mid+\mid \mathbb{X}f_y - \mu \tau_y \mid
		\label{tau-min}
		\end{equation}
		and 
		\begin{equation}
		\tau_z^{\max} =
		\mu(\mathbb{X}+\mathbb{Y})f_z-\mid \mathbb{Y}f_x + \mu \tau_x \mid-\mid \mathbb{X}f_y + \mu \tau_y \mid
		\label{tau-max}
		\end{equation}
		The equations~\eqref{boundaries} are combined in the following form for feet contacts:
		\begin{equation}
		\pm \begin{bmatrix}
		f_x\\f_y\\f_z\\ \tau_x \\ \tau_y 
		\end{bmatrix}_{\text{rf},\text{lf}} \leqslant 
		\begin{bmatrix}
		\mu f_z\\ \mu f_z \\f_z^{\max}\\ \mathbb{Y}f_z \\ \mathbb{X}f_z 
		\end{bmatrix}_{\text{rf},\text{lf}} \label{inequality1}
		\end{equation}
		
		Furthermore, for sliding contacts, we consider torques similarly to~\eqref{inequality1} in all directions inside the inequality constraints. Because, sliding contact forces have been considered in equality constraints to create the sliding motion. Note that, there is no $\tau_z$ element inside the vectors and it will be considered separately. We need to generate vectors in a way that we can apply it as an inequality constraint to the system. For this reason, we re-write these equations using $\bfG \in \mathbb{R}^{6 \times 21} $ and $\textbf{h} \in \mathbb{R}^{6} $ matrices and vectors:
		\begin{equation}
		\bfG_{\text{rf},\text{lf},\text{rh}}^1 \bfY \leqslant \textbf{h}_{\text{rf},\text{lf},\text{rh}}^1
		\end{equation}
		\begin{equation}
		\bfG_{\text{rf},\text{lf},\text{rh}}^2 \bfY \leqslant \textbf{h}_{\text{rf},\text{lf},\text{rh}}^2
		\end{equation}
		where:
		\begin{subequations}\label{G_matrices}
			\begin{align} 
			\bfG_{\text{rf}}^1 &= \begin{bmatrix}
			\textbf{0}_{6 \times 3} &
			\bfUpsilon_{\text{rf}}^1 &
			\textbf{0}_{6 \times 12} 
			\end{bmatrix} \\
			\bfG_{\text{rf}}^2 &= \begin{bmatrix}
			\textbf{0}_{6 \times 3} &
			\bfUpsilon_{\text{rf}}^2 &
			\textbf{0}_{6 \times 12} 
			\end{bmatrix} \\
			\bfG_{\text{lf}}^1 &= \begin{bmatrix}
			\textbf{0}_{6 \times 9} &
			\bfUpsilon_{\text{lf}}^1 &
			\textbf{0}_{6 \times 6} 
			\end{bmatrix} \\
			\bfG_{\text{lf}}^2 &= \begin{bmatrix}
			\textbf{0}_{6 \times 9} &
			\bfUpsilon_{\text{lf}}^2 &
			\textbf{0}_{6 \times 6} 
			\end{bmatrix} \\
			\bfG_{\text{rh}}^1 &= \begin{bmatrix}
			\textbf{0}_{6 \times 15} &
			\bfUpsilon_{\text{rh}}^1
			\end{bmatrix} \\
			\bfG_{\text{rh}}^2 &= \begin{bmatrix}
			\textbf{0}_{6 \times 15} &
			\bfUpsilon_{\text{rh}}^2
			\end{bmatrix}
			\end{align}
		\end{subequations}
		where
		$\bfUpsilon_{\text{rf},\text{lf}}^1 \in \mathbb{R}^{6 \times 6}$ is defined as:
		\begin{equation*}
		\bfUpsilon_{\text{rf},\text{lf}}^1 = \begin{bmatrix}
		1 & 0 & -\mu & 0 & \hdots \\
		0 & 1 & -\mu & \vdots \\
		\vdots & 0 & 1 & \\
		&  & -\mathbb{Y}_{\text{rf},\text{lf}} & 1 \\
		& & -\mathbb{X}_{\text{rf},\text{lf}} & & 1 & \vdots\\
		0& & 0 & &\hdots & 0
		\end{bmatrix}
		\end{equation*}
		
		Note that $\bfUpsilon_{\text{rf},\text{lf}}^2$ is same as $\bfUpsilon_{\text{rh},\text{lf}}^1$ except the diagonal values of the matrix which should be multiplied by $-1$. Also, for the sliding contact, we have:
		\begin{equation*}
		\bfUpsilon_{\text{rh}}^1 = \begin{bmatrix}
		0 & 0 & \hdots & & & 0\\
		0 &  &  &  \\
		\vdots &  & \ddots& & \vdots & \\
		0&  &  & 0 & 0 &\vdots\\
		\mathbb{X}_{\text{rh}}&0 & \hdots & 0& 1 & 0\\
		\mathbb{Y}_{\text{rh}}& 0& \hdots & &0 & 1
		\end{bmatrix}
		\end{equation*}
		and for $\bfUpsilon_{\text{rh}}^2$, just replace two last elemets on the diagonal of the matrix with -1.
		For the other side of the inequality, $h$ is zero vector for sliding contacts. For fixed contacts, we have:
		\begin{subequations}
			\begin{align}
			\textbf{h}_{\text{rf},\text{lf}}^1 &= \begin{bmatrix}
			0& 0& f_z^{\max}& 0& 0& 0 
			\end{bmatrix}^T \\
			\textbf{h}_{\text{rf},\text{lf}}^2 &= \begin{bmatrix}
			0& 0& f_z^{\min}& 0& 0& 0 
			\end{bmatrix}^T \\
			\textbf{h}_{\text{rh}}^{1,2} &= \textbf{0}_{6 \times 1}
			\end{align}
		\end{subequations}
		
		Besides, we should consider inequality constraints on $ \tau_z $. According to~\eqref{boundaries}, we divide it to two types of boundaries which should be implemented for all contacts:
		\begin{align}
		\tau_z -\tau_z^{\max} \leqslant 0  &\label{tawZ1} \\
		-\tau_z +\tau_z^{\min} \leqslant 0   &\label{tawZ2}
		\end{align} consider that there is two absolute values inside each of $\tau_z^{\max}$ and $\tau_z^{\min}$ based on~\eqref{tau-min} and~\eqref{tau-max} that each equation results in four more rows inside the inequality matrix that should be multiplied by $\bfY$. We introduce $\textbf{Gz} \in \mathbb{R}^{4 \times 21}$ matrices which covers the inequality constraints on $ \tau _ z$ element of wrenches:
		\begin{equation}
		\textbf{Gz}_{\text{rf},\text{lf},\text{rh}}^{1,2} \bfY\leqslant \textbf{0}_{4 \times 1}
		\end{equation}
		where:
		\begin{subequations}\label{Ztaw}
			\begin{align} 
			\textbf{Gz}_{\text{rf}}^1 &= \begin{bmatrix}
			\textbf{0}_{4 \times 3} &
			\bfPsi_{\text{rf}}^1 &
			\textbf{0}_{4 \times 12} 
			\end{bmatrix} \\
			\textbf{Gz}_{\text{rf}}^2 &= \begin{bmatrix}
			\textbf{0}_{4 \times 3} &
			\bfPsi_{\text{rf}}^2 &
			\textbf{0}_{4 \times 12} 
			\end{bmatrix}\\
			\textbf{Gz}_{\text{lf}}^1 &= \begin{bmatrix}
			\textbf{0}_{4 \times 9} &
			\bfPsi_{\text{lf}}^1 &
			\textbf{0}_{4 \times 6} 
			\end{bmatrix}\\
			\textbf{Gz}_{\text{lf}}^2 &= \begin{bmatrix}
			\textbf{0}_{4 \times 9} &
			\bfPsi_{\text{lf}}^2 &
			\textbf{0}_{4 \times 6} 
			\end{bmatrix}\\
			\textbf{Gz}_{\text{rh}}^1 &= \begin{bmatrix}
			\textbf{0}_{4 \times 15} &
			\bfPsi_{\text{rh}}^1
			\end{bmatrix}\\
			\textbf{Gz}_{\text{rh}}^2 &= \begin{bmatrix}
			\textbf{0}_{4 \times 15} &
			\bfPsi_{\text{rh}}^2 
			\end{bmatrix}
			\end{align}
		\end{subequations} where the $\bfPsi \in \mathbb{R}^{4 \times 6}$ matrices are different for each contact and should be defined separately. They are calculated as follows.
		Consider~\eqref{tawZ1} for the right foot. By implementing the amount of $\tau_z^{\max}$ from~\eqref{tau-max} and considering positive sign for the both of the absolute values, we get to the following equation:
		$$
		\mathbb{Y}_{\text{rh}}f_x + \mathbb{X}_{\text{rh}}f_y +\mathcal{C}_{\text{rf}}f_z+ \mu \tau_x +\mu \tau_y + \tau_z \leqslant 0
		$$
		where $\mathcal{C}_{\text{rf}} = -\mu(\mathbb{X}+\mathbb{Y})$ and is computed in the same way for other contacts.By considering negative sign for absolute values, the other three equations are available and $\bfPsi$ matrix for for upper bound of the right foot is calculated:
		\begin{equation*}
		\bfPsi_{\text{rf}}^1 = \begin{bmatrix}
		\mathbb{Y}_{\text{rf}}  & \mathbb{X}_{\text{rf}}  & \mathcal{C}_{\text{rf}} & \mu  & \mu & 1 \\
		-\mathbb{Y}_{\text{rf}} & \mathbb{X}_{\text{rf}}  & \mathcal{C}_{\text{rf}} & -\mu & \mu & 1 \\
		\mathbb{Y}_{\text{rf}}  & -\mathbb{X}_{\text{rf}} & \mathcal{C}_{\text{rf}} & \mu  & \mu & 1 \\
		-\mathbb{Y}_{\text{rf}} & -\mathbb{X}_{\text{rf}} & \mathcal{C}_{\text{rf}} & -\mu & \mu & 1 
		\end{bmatrix}
		\end{equation*}
		
		Notice that the first row of this matrix corresponds to the above equation. Other matrices for the lower and upper bound of the contacts will be calculated in the same way.

		\subsection{QP Formulation} \label{QP_formualation}

		The goal of centroidal QP is to achieve $\bfY_{\text{des}}$ as desired amount of these decision variables as much as possible. Therefore, the minimization problem is:
		\begin{equation}
		\lVert \bfY - \bfY_{\text{des}} \rVert _2  \label{Ydes}
		\end{equation} 
		where the desired position for CoM is the `middle' of the CSA and for feet contacts, desired quantity is equal to the wrench distribution for both. 
		
		According to~\eqref{Ydes} and sliding and non-sliding conditions, the QP formulation is in the following form:
		\begin{subequations}\label{QPformulation}
			\begin{align}
			\min_\bfY \ &{\frac{1}{2}\bfY^T\bfP\bfY + \textbf{q}^T\bfY} \\
			&\bfG\bfY \leqslant \textbf{h} \\
			&\bfA\bfY = \textbf{b}
			\end{align}
		\end{subequations} where $\bfP = 2 \bfI_{21 \times 21}$ and $\bfq = -2\bfY_{des}$. By introducing $\bfG_{\text{rf}}^T$ as transpose matrix of $\bfG_{\text{rf}}$ and the same for the other contacts, matrices and vectors of the equality constraints in~\eqref{QPformulation} are:
		\begin{subequations} 
			\begin{align}
			\bfG &= \begin{bmatrix}
			\bfG_{\text{rf}}^{1,T} & \bfG_{\text{rf}}^{2,T} &
			\bfG_{\text{lf}}^{1,T} & \bfG_{\text{lf}}^{2,T} &
			\bfG_{\text{rh}}^{1,T} & \bfG_{\text{rh}}^{2,T}
			\end{bmatrix}_{21 \times 60}^T \\
			\textbf{h} &= \begin{bmatrix}
			\textbf{h}_{\text{rf}}^{1,T} & \textbf{h}_{\text{rf}}^{2,T} &
			\textbf{h}_{\text{lf}}^{1,T} & \textbf{h}_{\text{lf}}^{2,T} &
			\textbf{h}_{\text{rh}}^{1,T} & \textbf{h}_{\text{rh}}^{2,T}
			\end{bmatrix}_{21 \times 1}^T
			\end{align}
		\end{subequations}
		On the other hand, for equality constraints,~\eqref{w} from section~\ref{wrenches-section} and~\eqref{sl_final} from section~\ref{sliding-section}, should be used. To combine these constraints in one equation and be able to use them in QP directly, matrix $\bfA \in \mathbb{R}^{12\times 21}$ and vector $ \bfb \in \mathbb{R}^{12}$  are defined as follows:
		\begin{subequations}
			\begin{align}
			\bfA &=
			\begin{bmatrix}
			\bfE_{\text{m1}} & \bfE_{\text{rf}} & \bfE_{\text{lf}} & \bfE_{\text{rh}}\\
			0_{6\times 3} & 0_{6\times 6} & 0_{6\times 6} & \bfS^1 - \bfS^2
			\end{bmatrix}\\
			\textbf{b} &=
			\begin{bmatrix}
			-\bfE_\text{m2}^\text{T} \in \mathbb{R}^{6 \times 1}& &-\bff_x &0& \hdots&0
			\end{bmatrix}^\text{T}
			\end{align}
		\end{subequations}
		
		\subsection{Controller Specification} 
		Position of the CoM and wrenches of contacts is calculated by centroidal QP. These values should be applied on the real robot as commands. For this purpose, we use a whole-body dynamic controller based on another QP formulation introduced in~\cite{Bouyarmane2017} using three following tasks.
		
		\subsubsection{Posture task} is a regularization task based on degrees of freedom $q$ that brings the robot to a reference joint-angle half-sitting configuration $\bfq_\text{half-sit}$ and its derivatives by considering task stiffness $\bfK$ via:
		\begin{align}
		\bfqdd & = \bfK (\bfq_\text{half-sit} - \bfq) - 2 \sqrt{\bfK} \bfqd
		\end{align}
		
		\subsubsection{End-effector admittance task} takes a desired position $\bfp^d \in \mathbb{R}^6$ as target in world frame, desired wrench $\bfw^d \in \mathbb{R}^6$ in sensor frame and admittance gains $\mathcal{A} \in \mathbb{R}^6$. For a given degree of freedom~$i$, the task is a position task if $\mathcal{A}_i = 0$ and a force task if $\mathcal{A}_i \neq 0$ and in this case, force feedback is applied by $\dot{\bfp}_i = \mathcal{A}_i (\bfw_i^d - \bfw_i)$. In experiments, we apply this task to the right hand of the robot which is in contact with the wall.
		
		\subsubsection{CoM task} takes as target a desired position $\bfc^d \in \mathbb{R}^3$, velocity $\bfcd^d \in \mathbb{R}^3$ and acceleration $\bfcdd^d \in \mathbb{R}^3$ in world frame. This is a standard second-order task. Internally, it will realize:
		\begin{align}
		\bfcdd & = K (\bfc^d - \bfc) + B (\bfcd^d - \bfcd) + \bfcdd^d
		\end{align}
		where $B$ is the task damping (usually $2 \sqrt{K}$) and $\bfcd = \bfJ_\text{com} \bfqd$ is the CoM velocity where $\bfJ$ shows Jacobian matrix.
		
		Finally, we regulate the weight distribution between the two feet on the ground while the end effector of the robot acts on the contact surface. For this purpose, we use:
		
		\subsubsection{Foot force difference control~\cite{kajita2010iros}} applies to two end-effectors whose targets are in the same plane, in our case the left foot and right foot. The force difference $(f_{z,\text{lf}} - f_{z,\text{rf}})$ is regulated to the value $(f^d_{z,\text{lf}} - f^d_{z,\text{rf}})$ provided by the centroidal QP by applying damping control to a virtual offset $z$:
		\begin{align}
		\dot{z} & = A_z ((f_{z,\text{lf}} - f_{z,\text{rf}}) - (f^d_{z,\text{lf}} - f^d_{z,\text{rf}}))
		\end{align}
		The velocity $-\dot{z}$ is then applied to the left foot, while $+\dot{z}$ is applied to the right foot.
		
		\section{Experiments and Results} \label{exp}
		
		We implemented our new methodology for keeping the balance and did several experiments with HRP-4 humanoid robot. In this section, we discuss three of these experiments and show the performance of our controller in pushing and wiping scenarios. The baseline for our study is considering a fixed position for the robot's CoM. The Figure~\eqref{failed} shows the robot scrabbling to increase the normal force on the hand contact and achieve the target force of $60$~N but fails. 
		\begin{figure}[!tb]
			\centering
			\includegraphics[width=0.4\textwidth, height=0.3\textwidth]{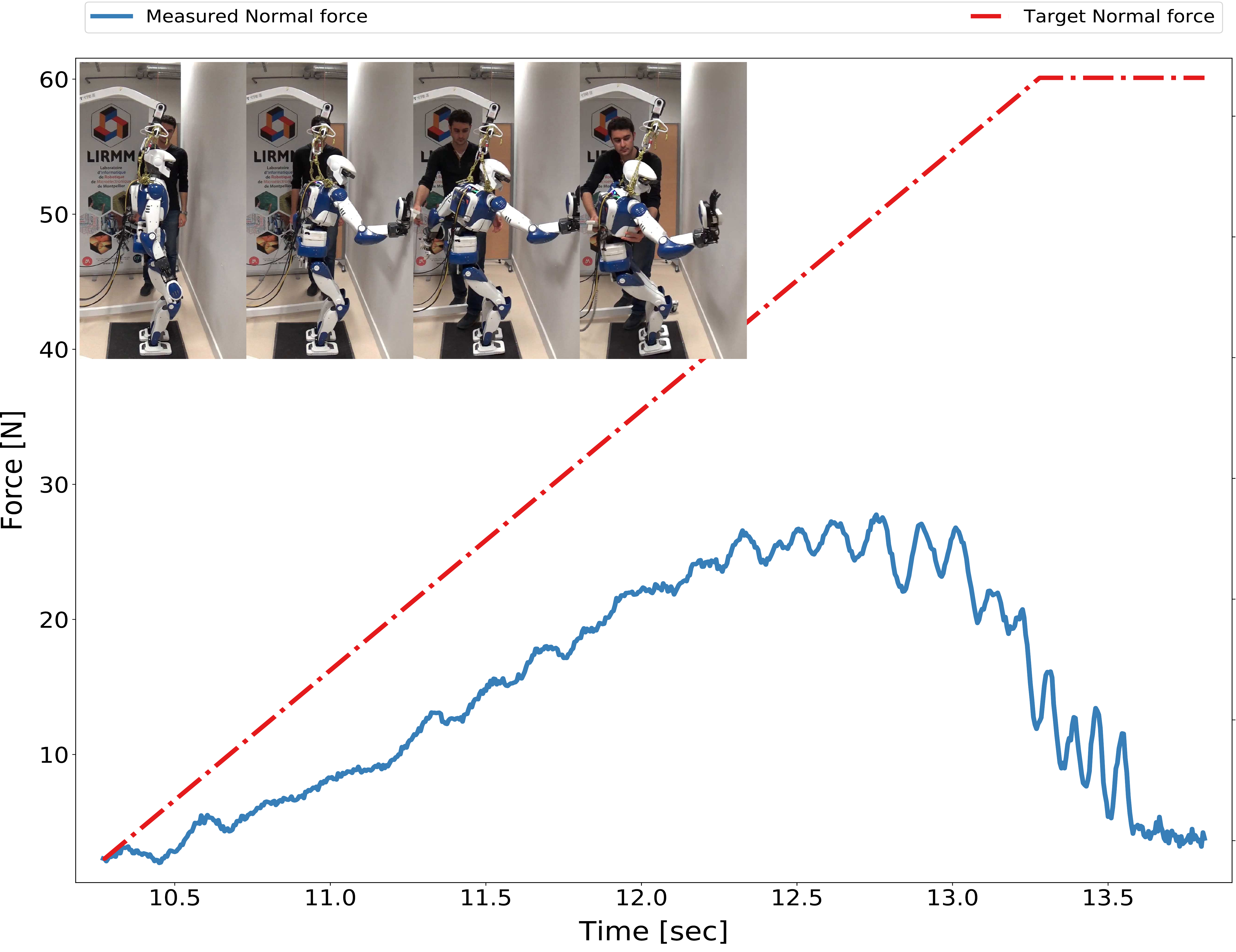}
			\caption{Normal force tracking while Pushing the wall using fixed CoM.}
			\label{failed}
		\end{figure}
		
		Force tracking of admittance control shows that the failure occurs in less than $20$~N normal force and the robot is not able to achieve the target force with this posture anymore.
		On the other hand, we did the same experiment with our proposed controller. The Figure~\eqref{success} shows successful tracking of the normal force with the generated posture of the robot due to the position of CoM and kinematics of the robot.
		\begin{figure}[!tb]
			\centering
			\includegraphics[width=0.4\textwidth, height=0.3\textwidth]{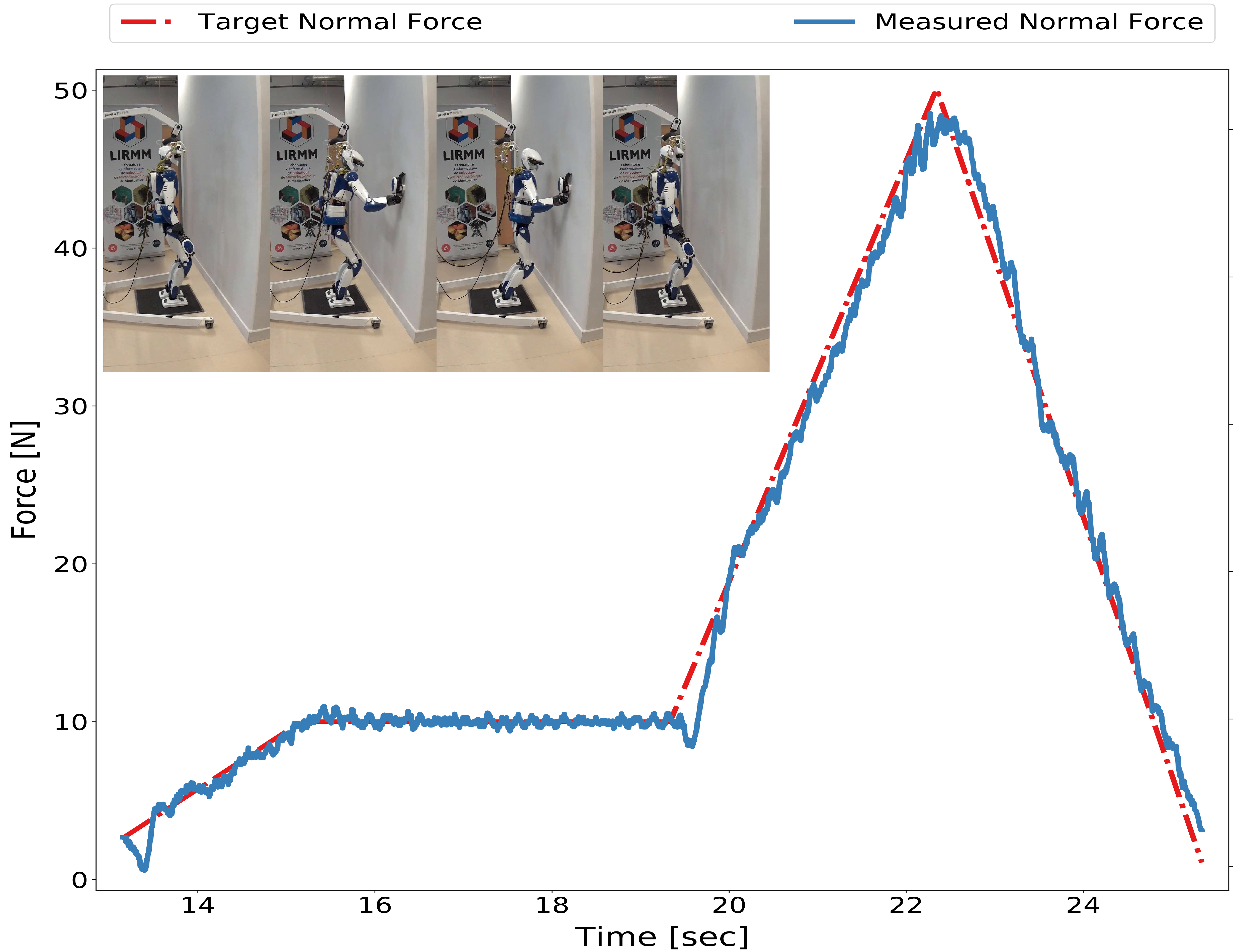}
			\caption{Normal force tracking while Pushing the wall using proposed strategy}
			\label{success}
		\end{figure}
		
		Also, the trajectory of CoM while increasing and decreasing the right-hand force is shown in Figure~\eqref{CoM}. Position of the CoM moves forward by increasing the target normal force and lays back by decreasing the force.
		\begin{figure}[!tb]
			\centering
			\includegraphics[width=0.4\textwidth, height=0.3\textwidth]{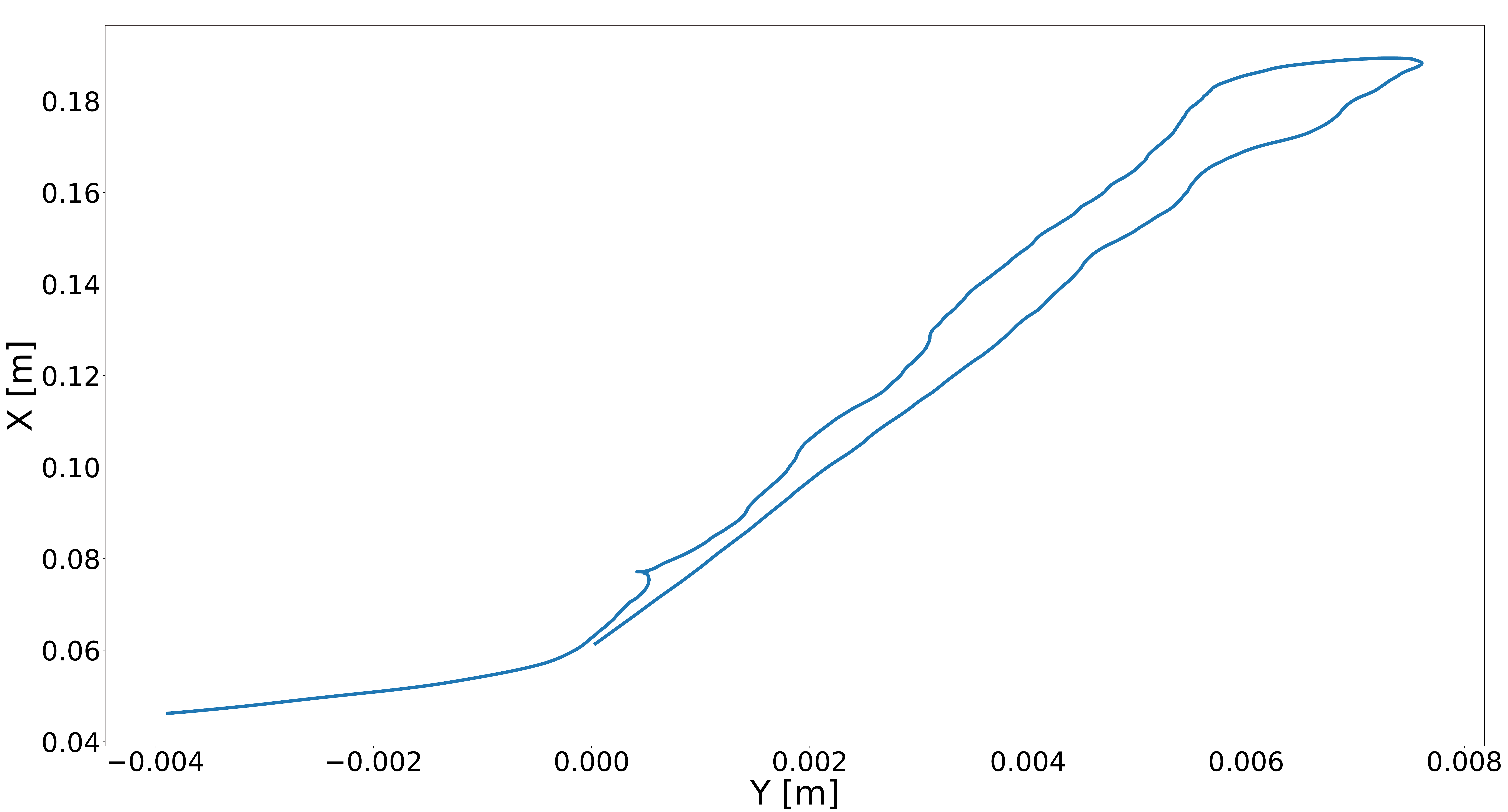}
			\caption{Position of CoM while pushing the wall.}
			\label{CoM}
		\end{figure}
		The last experiment deals with sliding contact and shows the performance of the controller while wiping a vertical surface with a normal force of $30$~N on sliding contact. Figure~\eqref{wiping} shows the normal force tracking of sliding contact.
		\begin{figure}[!tb]
			\centering
			\includegraphics[width=0.4\textwidth, height=0.3\textwidth]{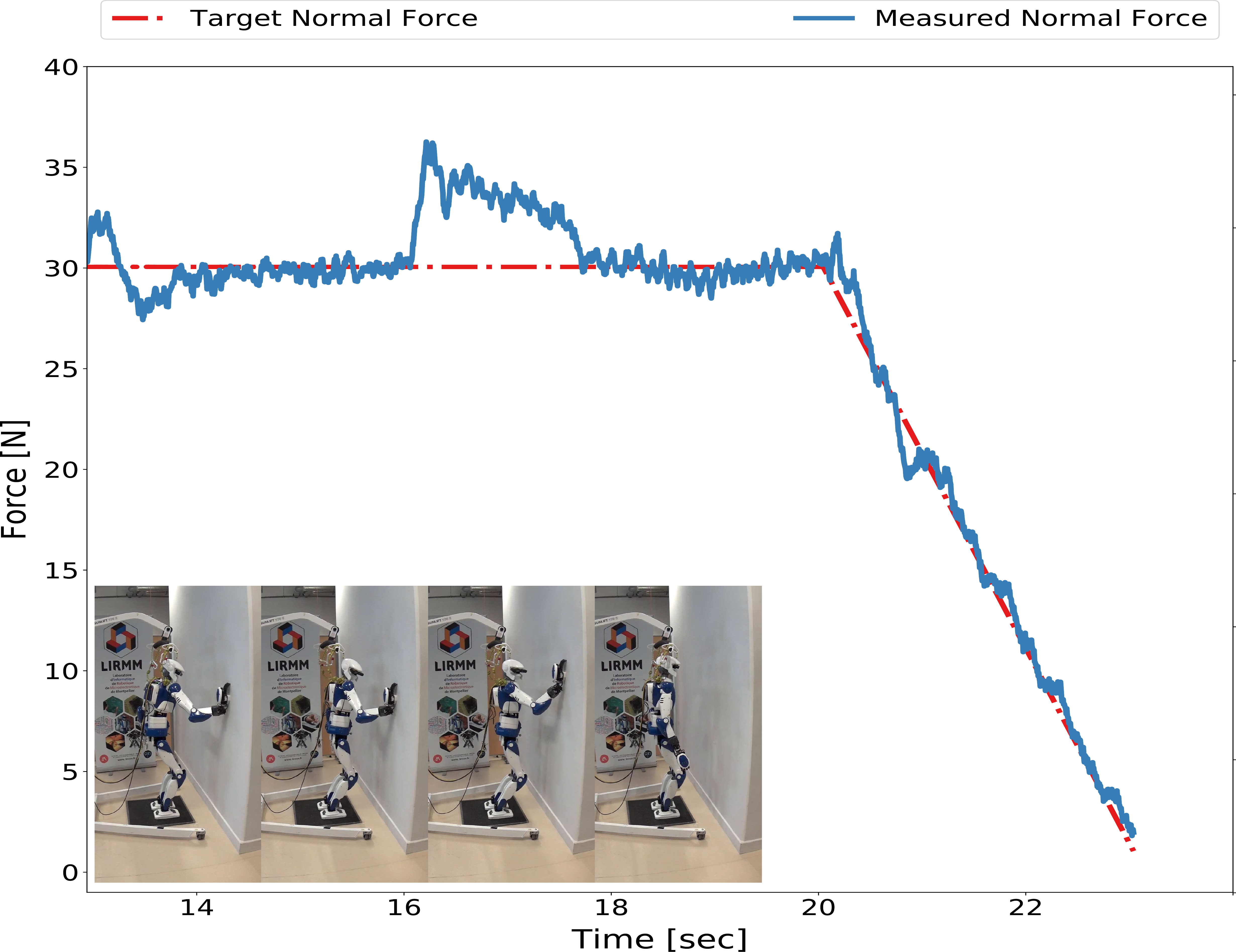}
			\caption{Normal force tracking while wiping the wall using proposed strategy.}
			\label{wiping}
		\end{figure}
		In this experiment, the normal force is set to $30$~N} by admittance task of the controller and starts wiping from $16^{th}$ second. This is the reason of error occurred at this moment but then the normal force converges to the target value. Experiments are available in accompanying video \footnote{\url{https://www.youtube.com/watch?v=Wai-Lp4e5FE}}.
	
	\section{Conclusion} \label{conc}
	
	Stability criteria for humanoid robots based on the position of the CoM have been studied in the current work. A methodology for constructing CSA in multi-contact conditions is introduced. This method, covers multiple fixed and sliding contacts. For setting CoM inside this the CSA, we use centroidal QP by considering constraints. The robot achieves to the desired configuration by implementing inverse kinematics of the system according to the position of the CoM.
	
	Simulations and experiments show that proposed balance control is valid for both fixed and sliding contacts in practice. Also, by using this method, the robot is able to achieve a proper body configuration in order to reach target forces in contacts. Online estimation of the contact friction should be addressed as part of future work. The transition between fixed and sliding contact modes, which is still a challenge to solve, causes errors in target force tracking. 
	
	\bibliographystyle{ieeetr}
	\bibliography{refs}
	
	\begin{comment}
	\begin{equation*}
	\Psi_{\text{rf}}^2 = \begin{bmatrix}
	\mathbb{Y}_{\text{rf}}  & \mathbb{X}_{\text{rf}}  & \mathcal{C}_{\text{rf}} & -\mu  & -\mu & -1 \\
	-\mathbb{Y}_{\text{rf}} & \mathbb{X}_{\text{rf}}  & \mathcal{C}_{\text{rf}} & \mu & -\mu & -1 \\
	\mathbb{Y}_{\text{rf}}  & -\mathbb{X}_{\text{rf}} & \mathcal{C}_{\text{rf}} & -\mu  & \mu & -1 \\
	-\mathbb{Y}_{\text{rf}} & -\mathbb{X}_{\text{rf}} & \mathcal{C}_{\text{rf}} & \mu & \mu & -1 
	\end{bmatrix}
	\end{equation*}
	
	\begin{equation*}
	\Psi_{\text{lf}}^1 = \begin{bmatrix}
	\mathbb{Y}_{\text{lf}}  & \mathbb{X}_{\text{lf}}  & \mathcal{C}_{\text{lf}} & \mu  & \mu & 1 \\
	-\mathbb{Y}_{\text{lf}} & \mathbb{X}_{\text{lf}}  & \mathcal{C}_{\text{lf}} & -\mu & \mu & 1 \\
	\mathbb{Y}_{\text{lf}}  & -\mathbb{X}_{\text{lf}} & \mathcal{C}_{\text{lf}} & \mu  & -\mu & 1 \\
	-\mathbb{Y}_{\text{lf}} & -\mathbb{X}_{\text{lf}} & \mathcal{C}_{\text{lf}} & -\mu & -\mu & 1 
	\end{bmatrix}
	\end{equation*}
	
	\begin{equation*}
	\Psi_{\text{lf}}^2 = \begin{bmatrix}
	\mathbb{Y}_{\text{lf}}  & \mathbb{X}_{\text{lf}}  & \mathcal{C}_{\text{lf}} & -\mu  & -\mu & -1 \\
	-\mathbb{Y}_{\text{lf}} & \mathbb{X}_{\text{lf}}  & \mathcal{C}_{\text{lf}} & \mu & -\mu & -1 \\
	\mathbb{Y}_{\text{lf}}  & -\mathbb{X}_{\text{lf}} & \mathcal{C}_{\text{lf}} & -\mu  & \mu & -1 \\
	-\mathbb{Y}_{\text{lf}} & -\mathbb{X}_{\text{lf}} & \mathcal{C}_{\text{lf}} & \mu & \mu & -1 
	\end{bmatrix}
	\end{equation*}
	
	\begin{equation*}
	\Psi_{\text{rh}}^1 = \begin{bmatrix}
	-\mathcal{C}_{\text{rh}}  & \mathbb{X}_{\text{rh}}   & \mathbb{Y}_{\text{rh}} & 1 & -\mu  & -\mu \\
	-\mathcal{C}_{\text{rh}}  & \mathbb{X}_{\text{rh}}   & -\mathbb{Y}_{\text{rh}} & 1 & -\mu  & \mu \\
	-\mathcal{C}_{\text{rh}}  & -\mathbb{X}_{\text{rh}}  & \mathbb{Y}_{\text{rh}} & 1 & \mu  & -\mu \\
	-\mathcal{C}_{\text{rh}}  & -\mathbb{X}_{\text{rh}}  & -\mathbb{Y}_{\text{rh}} & 1 & \mu  & \mu 
	\end{bmatrix}
	\end{equation*}
	
	\begin{equation*}
	\Psi_{\text{rh}}^1 = \begin{bmatrix}
	\mathcal{C}_{\text{rh}}  & -\mathbb{X}_{\text{rh}}   & -\mathbb{Y}_{\text{rh}} & 1 & -\mu  & -\mu \\
	\mathcal{C}_{\text{rh}}  & -\mathbb{X}_{\text{rh}}   & \mathbb{Y}_{\text{rh}} & 1 & -\mu  & \mu \\
	\mathcal{C}_{\text{rh}}  & \mathbb{X}_{\text{rh}}  & -\mathbb{Y}_{\text{rh}} & 1 & \mu  & -\mu \\
	\mathcal{C}_{\text{rh}}  & \mathbb{X}_{\text{rh}}  & \mathbb{Y}_{\text{rh}} & 1 & \mu  & \mu 
	\end{bmatrix}
	\end{equation*}
	and $\mathcal{C}$ is defined as:
	$$\mathcal{C}_{\text{rf},lf,rh} = -\mu(\mathbb{X}_{\text{rf},lf,rh}+\mathbb{Y}_{\text{rf},lf,rh}) $$
	\end{document}